\documentclass{article} %
\usepackage{arxiv,times}

\usepackage{amsmath,amsfonts,bm}

\def\eqref#1{equation~\ref{#1}}

\def\1{\bm{1}}

\DeclareMathAlphabet{\mathsfit}{\encodingdefault}{\sfdefault}{m}{sl}
\SetMathAlphabet{\mathsfit}{bold}{\encodingdefault}{\sfdefault}{bx}{n}

\DeclareMathOperator*{\argmax}{arg\,max}

\usepackage{hyperref}
\usepackage{url}
\usepackage[noabbrev,capitalise]{cleveref}
\usepackage[english]{babel}
\usepackage{amsthm}
\usepackage[]{caption}
\usepackage{tabu}
\usepackage{subcaption}
\usepackage{wrapfig}
\usepackage{multirow}
\usepackage{multirow,makecell}
\usepackage{booktabs,siunitx}
\usepackage{graphicx}
\usepackage{amssymb}
\usepackage{pifont}%

\newtheorem{theorem}{Theorem}[section]
\newtheorem{condition}[theorem]{Condition}

\DeclareCaptionFont{blue}{\color{blue}}
\DeclareRobustCommand\onedot{\futurelet\@let@token\@onedot}
\def\onedot{. } %
\def\eg{\emph{e.g}\onedot} 
 
\newcommand{\vbar}{\, | \,}

\newcommand{\cmark}{\ding{51}}%
\newcommand{\xmark}{\ding{55}}%

\title{Meta Temporal Point Processes}

\author{Wonho Bae\\
University of British Columbia \& Borealis AI\\
\texttt{whbae@cs.ubc.ca}
\And
Mohamed Osama Ahmed\\
Borealis AI\\
\texttt{mohamed.o.ahmed@borealisai.com}
\And
Frederick Tung\\
Borealis AI\\
\texttt{frederick.tung@borealisai.com\:\:\:\:\:\,
}
\And
Gabriel L. Oliveira\\
Borealis AI\\
\texttt{gabriel.oliveira@borealisai.com}
}

\newcommand\mohamed[1]{{\color{black} #1}}

\iclrfinalcopy %
\begin{document}

\maketitle

\begin{abstract}

A temporal point process (TPP) is a stochastic process where its realization is a sequence of discrete events in time.
Recent work in TPPs model the process using a neural network in a supervised learning framework, where a training set is a collection of all the sequences.
In this work, we propose to train TPPs in a meta learning framework, where each sequence is treated as a different task, via a novel framing of TPPs as neural processes (NPs). 
We introduce context sets to model TPPs as an instantiation of NPs. Motivated by attentive NP, we also introduce local history matching to help learn more informative features.
We demonstrate the potential of the proposed method on popular public benchmark datasets and tasks, and compare with state-of-the-art TPP methods.
\end{abstract}

\section{Introduction}
With the advancement of deep learning, there has been growing interest in modeling temporal point processes (TPPs) using neural networks. 
Although the community has developed many innovations in how neural TPPs encode the history of past events~\citep{nf2021bilovs} or how they decode these representations into predictions of the next event~\citep{intensity_free2019shchur,gen_tpp2022lin},
the general training framework for TPPs has been \textit{supervised learning} where a model is trained on a collection of all the available sequences.
However, supervised learning is susceptible to overfitting, especially in high noise environments, and generalization to new tasks can be challenging.

In recent years, \textit{meta learning} has emerged as an alternative to supervised learning as it aims to adapt or generalize well on new tasks, which resembles how humans can learn new skills from a few examples.
Inspired by this, we propose to train TPPs in a meta learning framework.
To this end, we treat each sequence as a ``task'', since it is a realization of a stochastic process with its own characteristics.
For instance, consider the pickup times of taxis in a city. The dynamics of these event sequences are governed by many factors such as location, weather and the routine of a taxi driver, which implies the pattern of each sequence can be significantly different from each other.
Under the supervised learning framework, a trained model tends to capture the patterns seen in training sequences well, but it easily breaks on unseen patterns.

As the goal of modeling TPPs is to estimate the true probability distribution of the next event time given the previous event times, we employ Neural Processes (NPs), a family of the model-based meta learning with stochasticity, to explain TPPs.
In this work, we formulate neural TPPs as NPs by satisfying some conditions of NPs, and term it as Meta TPP.
Inspired by the literature in NP, we further propose the Meta TPP with a cross-attention module, which we refer to as Attentive TPP. 
We demonstrate the strong potential of the proposed method through extensive experiments. %

Our contributions can be summarized as follows,
\begin{itemize}
    \item \mohamed{ To the best of our knowledge, this is the first work that formulates the TPP problem in a meta learning framework, opening up a new research direction in neural TPPs.} 
    \item \mohamed{Inspired by the NP literature, we present a conditional meta TPP formulation, followed by a latent path extension, culminating with our proposed Attentive TPP model. }
    
    \item \mohamed{The experimental results show that our proposed Attentive TPP model achieves state-of-the-art results on four widely used TPP benchmark datasets, and is more successful in capturing periodic patterns on three additional datasets compared to previous methods.}

    \item \mohamed{We demonstrate that our meta learning TPP approach can be more robust in practical deployment scenarios such as noisy sequences and distribution drift.}
\end{itemize}

\section{Preliminaries}
\label{sec:preliminary}

\textbf{Neural processes.}
A general form of optimization objective in \textbf{supervised learning} is defined as,
\begin{align}
    \theta^* = \argmax_{\theta} \mathbb{E}_{B \sim p(\mathcal{D})} \left[ \sum_{(x,y) \in B}\log p_{\theta}(y \vbar x) \right]
    \label{eq:supervised_learning}
\end{align}
where $\mathcal{D} :=\{(x^{(i)}, y^{(i)})\}_{i=1}^{|\mathcal{D}|}$ for an input $x$ and label $y$, and $B$ denotes a mini-batch set of $(x,y)$ data pairs.
Here, the goal is to learn a model $f$ parameterized by $\theta$ that maps $x$ to $y$ as $f_\theta: x \rightarrow y$. 

In recent years, \textbf{meta learning} has emerged as an alternative  to supervised learning as it aims to adapt or generalize well on new tasks \citep{MANN2016santoro}, which resembles how humans learn new skills from few examples.
In meta learning, we define a meta dataset, a set of different tasks, as $\mathcal{M} := \{\mathcal{D}^{(i)}\}_{i=1}^{|\mathcal{M}|}$.
Here, $\mathcal{D}^{(i)}$ is a dataset of $i$-th task consisting of a context and target set as $\mathcal{D} := \mathcal{C} \cup \mathcal{T}$.
The objective of meta learning is then defined as,
\begin{align}
    \theta^* = \argmax_{\theta} \mathbb{E}_{B_\mathcal{D} \sim p(\mathcal{M})} \left[\sum_{(\mathcal{C},\mathcal{T}) \in B_\mathcal{D}} \log p_{\theta}(\mathcal{Y}_\mathcal{T} \vbar \mathcal{X}_\mathcal{T} , \mathcal{C}) \right]
    \label{eq:meta_learning}
\end{align}
where $B_\mathcal{D}$ denotes a mini-batch set of tasks. Also, $\mathcal{X}_\mathcal{T}$ and $\mathcal{Y}_\mathcal{T}$ represent inputs and labels of a target set, respectively.
Unlike supervised learning, the goal is to learn a mapping from $x$ to $y$ given $\mathcal{C}$: more formally, $f_\theta (\cdot, \mathcal{C}): x \rightarrow y$.
Although meta learning is a powerful framework to learn fast adaption to new tasks,
it does not provide uncertainty for its predictions, which is becoming more important in modern machine learning literature as a metric to measure the reliability of a model.

To take the uncertainty into account for meta learning, \textbf{Neural processes} (NPs) have been proposed \citep{np2018garnelo,cnp2018garnelo2018}.
Instead of finding point estimators as done in regular meta learning models, NPs learn a probability distribution of a label $y$ given an input $x$ and context set $\mathcal{C}$: $p_{\theta}(y | x, \mathcal{C})$.
In this work, we frame TPPs as meta learning instead of supervised learning, for the first time.
To this end, we employ NPs to incorporate the stochastic nature of TPPs.
In \cref{subsec:tpp_as_np}, we propose a simple modification of TPPs to connect them to NPs, which enables us to employ a rich line of works in NPs to TPPs as described in \cref{subsec:meta_tpp} and \cref{subsec:attentive_tpp}.

\begin{figure}[t!]
    \centering
    \includegraphics[width=0.96\textwidth]{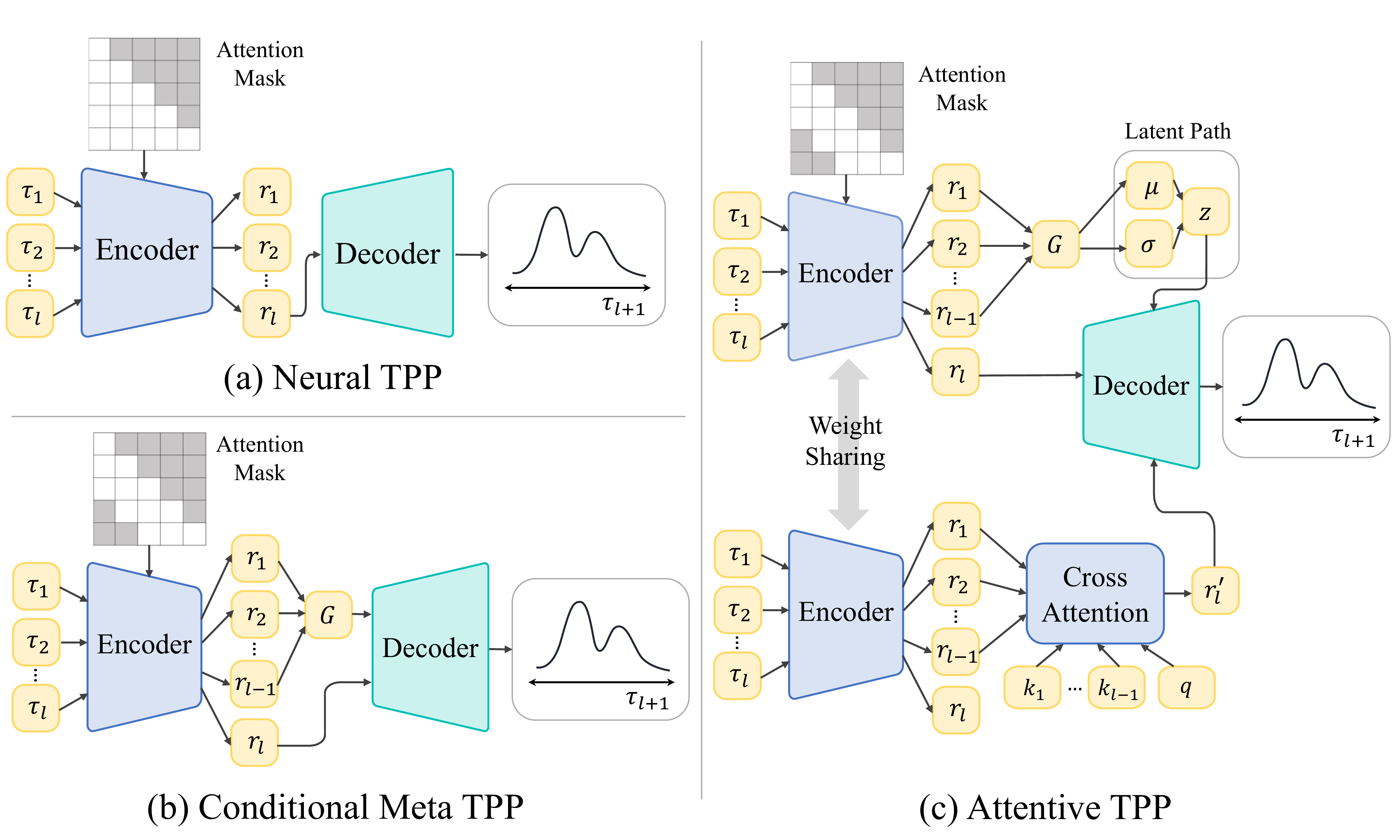}
    \caption{Overall architectures of TPP models.}
    \label{fig:overall}
    \vspace{-2mm}
\end{figure}

\textbf{Neural temporal point processes.} 
TPPs are stochastic processes where their realizations are sequences of discrete events in time.
In notations, a collection of event time sequences is defined as $\mathcal{D} :=\{s^{(i)}\}_{i=1}^{|\mathcal{D}|}$ where $s^{(i)} = (\tau^{(i)}_{1}, \tau^{(i)}_{2}, \dots, \tau^{(i)}_{L_i})$ and $L_i$ denotes the length of $i$-th sequence.
The history of studying TPPs started decades ago~\citep{tpp2007david}, but in this work, we focus on neural TPPs where TPPs are modeled using neural networks~\citep{review2021shchur}.
As described in \cref{fig:overall}a, a general form of neural TPPs consists of an encoder, which takes a sequence of previous event times and outputs a history embedding, and a decoder which takes the history embedding and outputs probability distribution of the time when the next event happens.

Previous works of neural TPPs are auto-regressively modeled in a supervised learning framework.
More formally, the objective of neural TPPs are defined as,
\begin{align}
    \theta^* &= \argmax_{\theta} \mathbb{E}_{B \sim p(\mathcal{D})} \left[ \sum_{i=l}^{|B|} \sum_{l=1}^{L_i-1} \log p_{\theta}(\tau^{(i)}_{l+1} \vbar \tau^{(i)}_{\leq l}) \right]
    \label{eq:tpp_supervised_learning}
\end{align}
where $B \sim p(\mathcal{D})$ denotes a mini-batch of event time sequences.
To frame TPPs as NPs, we need to define a target input and context set shown in \cref{eq:meta_learning}, from an event time history $\tau_{\leq l}$, which will be described in the following section.

\section{Meta Temporal Point Process and Its Variants}
\label{sec:meta_tpp}

\subsection{Temporal Point Processes as Neural Processes}
\label{subsec:tpp_as_np}

To frame TPPs as NPs, we treat each event time sequence $s$ as a task for meta learning, which intuitively makes sense since each sequence is a realization of a stochastic process.
For instance, the transaction times of different account holders are very different from each other due to many factors
including an account holder's financial status and characteristics.

With the new definition of tasks, we define a target input and context set for a conditional probability distribution of meta learning shown in \cref{eq:meta_learning}, using previous event times $\tau_{\leq l}$.
There are many ways to define them but a target input and context set need to be semantically aligned since the target input will be an element of the context set for the next event time prediction.
Hence, we define a target input for $\tau_{l+1}$ as the latest ``local history'' $\tau_{l-k+1 : l}$ where $k$ is the window size of the local history.
Similarly, a context set for $\tau_{l+1}$ is defined as $\mathcal{C}_{l} := \{\tau_{t-k+1:t } \}_{t=1}^{l-1}$.
Here, if $t - k \leq 0$, we include event times from $\tau_1$. 
With Transformer structure, it is easy to efficiently compute the feature embeddings for the context set $\mathcal{C}$.
\cref{fig:overall}b shows a schematic of the Conditional Meta TPP  with a mask (shaded) used for an example case of $5$ event times with a local history window size of $k=3$.
Then, the feature embedding $r_l$ contains information of $\tau_{l-k+1:l}$.
With the notations for target inputs and context sets, we propose the objective of TPPs in a meta learning framework as,
\begin{align}
    \theta^* &= \argmax_{\theta} \mathbb{E}_{B \sim p(\mathcal{D})} \left[ \sum_{i=l}^{|B|} \sum_{l=1}^{L_i-1} \log p_{\theta}(\tau^{(i)}_{l+1} \vbar \tau^{(i)}_{l-k+1 : l}, \mathcal{C}^{(i)}_{l}) \right].
    \label{eq:tpp_meta_learning}
\end{align}
Note that we have only one target label $\tau^{(i)}_{l+1}$ to predict per event unlike the general meta learning objective in \cref{eq:meta_learning} where usually $|\mathcal{T}| > 1$.
It is because TPP models in general are trained to predict the next event time.
Modeling TPPs to predict multiple future event times would be an interesting future work, but it is out of scope of this work.

\textbf{Requirements for neural processes.}
Let $\mathcal{X}_{\mathcal{T}} := \{x_i\}_{i=1}^{|\mathcal{T}|}$ and $\mathcal{Y}_{\mathcal{T}}:= \{y_i\}_{i=1}^{|\mathcal{T}|}$ be a set of target inputs and labels, respectively, and $\pi$ be an arbitrary permutation of a set.
To design NP models, it is required to satisfy the following two conditions.
\begin{condition}[Consistency over a target set] 
\label{cond:consistency}
A probability distribution $p_\theta$ is consistent if it is consistent under permutation:
$p_\theta(\mathcal{Y}_{\mathcal{T}} \vbar \mathcal{X}_{\mathcal{T}}, \mathcal{C}) = p_\theta(\pi(\mathcal{Y}_{\mathcal{T}}) \vbar \pi(\mathcal{X}_{\mathcal{T}}), \mathcal{C})$, and marginalization: $p_\theta(y_{1:m} \vbar \mathcal{X}_{\mathcal{T}}, \mathcal{C}) = \int p_\theta(y_{1:n} \vbar \mathcal{X}_{\mathcal{T}}, \mathcal{C}) \, dy_{m+1:n}\,$ for any positive integer $m < n$. 
\end{condition}

\begin{condition}[Permutation invariance over a context set]
\label{cond:perm_inv}
$p_\theta(\mathcal{Y}_{\mathcal{T}} \vbar \mathcal{X}_{\mathcal{T}}, \mathcal{C}) = p_\theta(\mathcal{Y}_{\mathcal{T}} \vbar \mathcal{X}_{\mathcal{T}}, \pi(\mathcal{C}))$ 
\end{condition}

According to Kolmogorov extension theorem~\citep{kolmogorov2013oksendal}, a collection of finite-dimensional distributions is defined as a stochastic process if condition 3.1 is satisfied.
In NP literature, condition 3.1 is satisfied through factorization: it assumes target labels are independent to each other given a target input and a context set $\mathcal{C}$, in other words, $p_\theta(\mathcal{Y}_{\mathcal{T}} \vbar \mathcal{X}_{\mathcal{T}}, \mathcal{C}) = \Pi_{i=1}^{|\mathcal{T}|} p_\theta(y_i \vbar x_i, x_{<i}, y_{<i}, \mathcal{C}) \approx \Pi_{i=1}^{|\mathcal{T}|} p_\theta(y_i \vbar x_i, \mathcal{C})$~\citep{npf2020dubois}.
This assumption can be unrealistic if target labels are strongly dependent to previous target inputs and labels even after context representations are observed.
It is, however, not necessary to assume factorization to make TPPs as NPs.
As previously mentioned, we only care about predicting the next event time, which means $|\mathcal{Y}_\mathcal{T}| = 1$.
When a set contains only one element, its permutation is always itself.
More formally, the consistency under permutation of Condition 3.1 in TPPs: $p_\theta (\tau_{l+1} \vbar \tau_{l-k+1:l}, \mathcal{C}_l) = p_\theta (\pi(\tau_{l+1}) \vbar \pi(\tau_{l-k+1:l}), \mathcal{C}_l)$, is satisfied since $\pi(\{ \tau_{l+1} \}) = \{ \tau_{l+1} \}$ and $\pi(\{ \tau_{l-k+1:l} \}) = \{ \tau_{l-k+1:l} \}$. 
Also, the marginalization under permutation in Condition 3.1 is satisfied as marginalization is not applicable for $p_\theta (\tau_{l+1} \vbar \tau_{l-k+1:l}, \mathcal{C}_l)$ since the target label set $\tau_{l+1}$ contains only one element.

Recall that NP models learn a probability distribution of a target label $p_\theta$ given a target input and context set.
For computational efficiency (to make inference $\mathcal{O}(|\mathcal{C}| + |\mathcal{T}|)$ time), the feature representation of $\mathcal{C}$ should be invariant to the size of the context set, for which Condition 3.2 is required.
To satisfy Condition 3.2, we average-pool all the context features $r_1, r_2, \dots r_{l-1}$ to generate the global feature for a task $G$ as shown in \cref{fig:overall}b, and term it as \textit{conditional Meta TPP} following the terminology used in the NP literature.
Each context feature $r_1, r_2, \cdots, r_{l-1}$ represents a feature from a transformer encoder such as Transformer Hawkes Processes (THP), that encodes the corresponding local history of the context set $\mathcal{C}_l$. 
For instance, $r_i$ contains information of $\tau_{i-k+1:i}$.
To make $r_i$ only encode the subset of previous event times $\tau_{i-k+1:i}$ (instead of the whole previous event times $\tau_{\leq i}$), we mask out events that are outside of the local history window using an attention mask as shown in Figure 1(b) and (c), which is different from a regular attention mask shown in Figure 1(a).
Using the permutation invariant feature $G$ not only satisfies Condition 3.2, but also lets the decoder approximate the probability distribution of a target label given both a target input and context set instead of just a target input.
Now that we satisfy both requirements with a new architectural design, we can treat TPPs as NPs.

\textbf{Implementation.}
It can be expensive to compute the individual context feature $r_t$ for all $1\leq t < l$, from each element of the context set $\tau_{t-k+1:t} \in \mathcal{C}_l$: the time complexity of computing all the context features for a sequence is $\mathcal{O}(L^2)$. %
Instead of passing each element of a context set, using the Transformer architecture~\citep{attention2017vaswani}, we can simply pass the event times to obtain all the context features at once, of which time complexity is $\mathcal{O}(kL)$ where $k$ is the window size of a local history.
To this end, we employ the THP as the encoder.
Please refer to \citet{thp2020zuo} for details.

\subsection{Meta Temporal Point Process}
\label{subsec:meta_tpp}

In the NP literature, NPs are generally modeled as latent variable models.
Instead of using the deterministic global feature $G$ as an input to the decoder~\citep{cnp2018garnelo2018}, a latent variable $z$ is sampled from a probability distribution \eg multi-variate Gaussian, using parameters inferred from the global feature $G$~\citep{np2018garnelo}. 
As it is intractable to compute the log-likelihood for a latent variable model, \textit{amortized variational inference} (VI) can be used to approximate inference.
In the setting of TPPs, the evidence lower bound (ELBO) of variational inference with an inference network $p_\theta(z \vbar \mathcal{C}_{L})$ can be derived as,
\begin{align}
    \log p_\theta(\tau_{l+1} \vbar \tau_{l-k+1:l}, \mathcal{C}_l) &= \log \int p_\theta(\tau_l \vbar \tau_{l-k+1:l}, z) p_\theta(z \vbar \mathcal{C}_l) dz \\
    &\geq \mathbb{E}_{z} \left[ \log p_\theta(\tau_{l+1} \vbar \tau_{l-k+1:l}, z) \right] - KL(p_\theta(z \vbar \mathcal{C}_L) \vbar p_\theta(z \vbar \mathcal{C}_l)) \\
    &\approx \frac{1}{N}\sum_{n=1}^N \log p_\theta(\tau_l \vbar \tau_{l-k+1:l}, z_n) - KL(p_\theta(z \vbar \mathcal{C}_L) \vbar p_\theta(z \vbar \mathcal{C}_l))
    \label{eq:npvi}
\end{align}
where $N$ denotes the number of samples of $z \sim p_\theta(z \vbar \mathcal{C}_L)$ for Monte-Carlo approximation.
Here, $p_\theta(z \vbar \mathcal{C}_L)$ is the posterior given the context at the last ($L$-th) event, which contains all the events of the sequence $s$ (it is accessible in training time).
Minimizing $KL$-divergence between $p_\theta(z \vbar \mathcal{C}_L)$ and $p_\theta(z \vbar \mathcal{C}_l)$ is to make the global latent variable $z$ inferred from $\mathcal{C}_l$ to be similar to the latent variable of a sequence $z$ from $\mathcal{C}_L$, in training time.
To sample $z$, we use the reparameterization trick as $z = \mu + \sigma \odot \epsilon$ where $\epsilon \sim \mathcal{N}(0, I)$ as described in the latent path of \cref{fig:overall}c.
In inference, we approximate evaluation metrics such as negative log-likelihood or root mean squared error using Monte-Carlo samples.
But, as we do not have access to $\mathcal{C}_L$ at $l$-th event when $l < L$, we use $z$ from $p_\theta(z \vbar \mathcal{C}_l)$.
The detailed description of the evaluation metrics are provided in \cref{app:sec:eval}.

An advantage of a latent variable model is that it captures stochasticity of functions, which can be particularly beneficial to model TPPs since TPPs are stochastic processes.
Experiments in \cref{subsec:ablation} demonstrate it indeed helps to model TPPs over the deterministic case.
In particular, it is robust to noises (\cref{subsec:experiment}).
We term the latent variable model as \textit{Meta TPP} throughout the paper.

\textbf{Discussion.}
The existing TPP models treat all event sequences as realization of the same process whereas the Meta TPP treats each sequence as a realization of a distinct stochastic process.
We achieve this by conditioning on the global latent feature $z$ that captures task-specific characteristics.
For $z$ to be task-specific, it has to be distinct for different sequences but similar throughout different events $l \in [1, L-1]$ within the same sequence.
It is natural for the global features to be distinct by sequence but we need further guidance to make the global feature shared across all the event times in a sequence.
Due to the permutation invariance constraint implemented in average-pooling, $z$ cannot be very different at different event time: adding some addition context feature $r_i$ will not change $G$ as well as $z$ much.
In addition, the KL-divergence between $p_\theta(z \vbar \mathcal{C}_L)$ and $p_\theta(z \vbar \mathcal{C}_l)$ further enhances the task-specific characteristics of $z$. 
We provide more detailed discussion in \cref{app:sec:global_feature}

\subsection{Attentive Temporal Point Process}
\label{subsec:attentive_tpp}
Early works in NPs suffered from the underfitting problem. 
To alleviate this, \citet{attnnp2019kim} proposed AttentiveNP, which explicitly attends the elements in a context set to obtain a better feature for target inputs.
Inspired by this, we add a cross-attention module that considers the similarity between the feature of a target input and previous event times as described in \cref{fig:overall}c.
Given the local history (context) features $r_1, r_2, \dots r_{l-1}$ at $l$-th time step, the key-query-value pairs $K \in \mathbb{R}^{l-1 \times D}, q \in \mathbb{R}^{1 \times D}$, and $V \in \mathbb{R}^{l-1 \times D}$ for the cross-attention, are computed using their corresponding projection weights $W_K \in \mathbb{R}^{D \times D}, \: W_Q \in \mathbb{R}^{D \times D}$ as, 
\begin{align}
    K = R\cdot W_K, \: q = {r_l}^T\cdot W_Q , \: V = R \: \text{ where } \: R = [r_1, r_2, \dots, r_{l-1} ]^T.
\end{align}
Here, K corresponds to $[k_1, k_2, \dots, k_{l-1}]^T$ in \cref{fig:overall}c. 
The feature of $i$-th attention head $h_i$ are then computed as follows,
\begin{align}
    h_i &= Softmax(q\cdot K^T / \sqrt{D})\cdot V.
\end{align}
With $W \in \mathbb{R}^{HD \times D}$ and some fully connected layers denoted as $FC$, $r'_l \in \mathbb{R}^{1 \times D}$ is computed as,
\begin{align}
    r'_l = FC(\, [h_1, h_2, \dots, h_H] \cdot W \,).
\end{align}
Finally, the decoder takes the concatenated feature of $z, r_l$, and $r'_l$ as an input to infer a distribution.

In the TPP setting, it is common that there are multiple periodic patterns in the underlying stochastic process.
The cross-attention module provides an inductive bias to a model that the repeating event subsequences should have similar features. 
Our experiments in \cref{subsec:experiment} demonstrate that the explicit attention helps to model TPPs in general, especially when there are periodic patterns.

\textbf{Decoder.}
The decoder takes the concatenated feature of the global latent feature $z$, target input feature $r_l$ that encodes $\tau_{l-k:l-1}$, and attention feature $r^{\prime}$ from the attention module.
For the Meta TPP (without the attention module), the decoder takes as input the concatenated feature of $z$ and $r_l$.
Here, $z, r_l$, and $r^{\prime}$ are all $D$-dimensional vectors.
The decoder consists of two fully connected layers, and the input and hidden dimension of the decoder layers are either $2D$ or $3D$ depending on whether we use the feature from the attention module $r^{\prime}$.

The decoder outputs the parameters of the probability distribution of the next event time or $p_\theta(\tau_{l+1} \vbar \tau_{l-k+1:l}, z_m)$.
Inspired by the intensity-free TPP (Shchur et al., 2020), we use a mixture of log-normal distributions to model the probability distribution.
Formally, for $l \in [1, L-1]$, $\tau_{l+1} \sim MixLogNorm(\bm{\mu}_{l+1}, \bm{\sigma}_{l+1}, \bm{\omega}_{l+1})$ where $\bm{\mu}_{l+1}$ are the mixture means, $\bm{\sigma}_{l+1}$ are the standard deviations, and $\bm{\omega}_{l+1}$ are the mixture weights.

\section{Related Work}
\textbf{Neural temporal point processes.} 
Neural temporal point processes (NTPPs) have been  proposed to capture complex dynamics of stochastic processes in time. 
They are derived from traditional temporal point processes \citep{hawkes1971spectra,isham1979self,tpp2007david}. 
Models based on RNNs are proposed by \citep{recurrent2016du} and \citep{hawkes2017mei} to improve NTPPs by constructing continuous-time RNNs. 
More recent works use Transformers to capture long-term dependency~\citep{tpp_data2019kumar, self2020zhang, thp2020zuo, transformer2021yang, mix_tpp2022zhang}.
\citep{fully2019omi, intensity_free2019shchur, Sharma2021} propose intensity-free NTPPs to directly model the conditional distribution of event times. 

\citet{fully2019omi} propose to model a cumulative intensity with a neural network. 
But, it suffers from problems that the probability density is not normalised and negative event times receives non-zero probabilities. 
Alternatively, \citet{intensity_free2019shchur} suggest modelling conditional probability density by log-normal mixtures.
Transformer-based models like \citet{thp2020zuo, self2020zhang} propose to leverages the self-attention mechanism to capture long-term dependencies.
Another class of TPP methods called Neural Flows~\citep{nf2021bilovs}, are proposed to model temporal dynamics with ordinary differential equations learned by neural networks.  
Unlike the previous TPP methods, we frame TPPs as meta learning (not supervised learning) for the first time.

\textbf{Neural processes.}
Meta learning is a learning framework that aims to adapt or generalize well on new tasks.
There are three approaches in meta learning: metric-based~\citep{siamese2015koch,matching2016vinyals,relation2018sung,proto2017snell}, model-based~\citep{MANN2016santoro,metanet2017munkhdalai,hierarchical2018Grant} and optimization-based~\citep{maml2017finn,probMAML2018Finn,reptile2018nichol}.
Neural processes (NPs) is the model-based meta learning with stochasticity.
\citet{cnp2018garnelo2018} propose a conditional neural process as a new formulation to approximate a stochastic process using neural network architecture.
It succeeds the advantage of Gaussian Processes (GPs) as it can estimate the uncertainty of its predictions, without having expensive inference time.
\citet{np2018garnelo} generalize a conditional neural process by adding latent variables, which are approximated using variational inference.
Although NPs can adapt to new tasks quickly without requiring much computation, it suffers from underfitting problem.
To alleviate it, \citet{attnnp2019kim} propose a cross-attention module, which explicitly attends the elements in the context set to obtain better representations for the elements in the target set.
As another way to address the underfitting problem, \citet{convnp2019gordon} propose a set convolutional layer under the assumption of translation equivariance of inputs and outputs, which is expanded to the latent variable counterpart in \citet{npml2020foong}.

Transformer NP~\citep{transformer2022nguyen} is the most relevant work to ours.
Although it also models event sequences, it focuses on modeling regular time series: discrete and regularly-spaced time inputs with corresponding label values.
TPPs are different as they are continuous and irregularly-spaced time sequences not necessarily with corresponding label values. 

\section{Experiments}
\label{sec:experiments}

\subsection{Experiment Setting}
\label{subsec:experiment_setting}

\textbf{Datasets.}
To compare the effectiveness of models, we conduct experiments on $4$ popular benchmark datasets -- Stack Overflow, Mooc, Reddit, and Wiki, and $3$ datasets with strong periodic patterns we introduce -- Sinusoidal wave, Uber, and NYC Taxi.
Please refer to \cref{app:sec:datasets} for details.

\textbf{Metrics.}
We use the root mean squared error (RMSE) as the main metric along with the negative log-likelihood (NLL) as a reference since NLL can go arbitrary low if probability density is placed mostly on the ground truth event time.
RMSE may not be a good metric, either, if one ignores stochastic components of TPPs and directly trains a baseline on the ground truth event times to obtain point estimations of event times~\citep{review2021shchur}.
We train all the methods on NLL and obtain RMSE in test time to not abuse RMSE scores, keeping stochastic components of TPPs.
For marked TPP datasets, we extend the proposed method to make class predictions, and report accuracy.
For details about marked cases, please refer to \cref{app:sec:marked_tpp}.

\textbf{Baselines.}
We use intensity-free TPP~\citep{intensity_free2019shchur}, Neural flow~\citep{nf2021bilovs}, and Transformer Hawkes Processes (THP)~\citep{thp2020zuo} as baselines.
For intensity-free TPP and neural flow, we add the survival time of the last event to NLL and fix some bugs specified in their public repositories.
THP and its variants are originally based on intensity: they predict intensities from which log-likelihood and expectation of event times are computed.
It is, however, computationally expensive to compute them as it requires to compute integrals: especially, to compute the expected event times, it requires to compute double integrals, which can be quite expensive and complex to compute even with thinning algorithms described in \citet{hawkes2017mei}.
To work around it without losing performance, we add the mixture of log-normal distribution proposed in \citep{intensity_free2019shchur} as the decoder, and we call it THP$^+$.
For fair comparison, we fix the number of parameters of the models in between $50$K and $60$K except the last fully-connected layer for class predictions since it depends on the number of classes. 

\begin{figure*}[t!]
    \begin{subfigure}[t]{0.48\textwidth}
        \centering
        \includegraphics[width=0.95\textwidth]{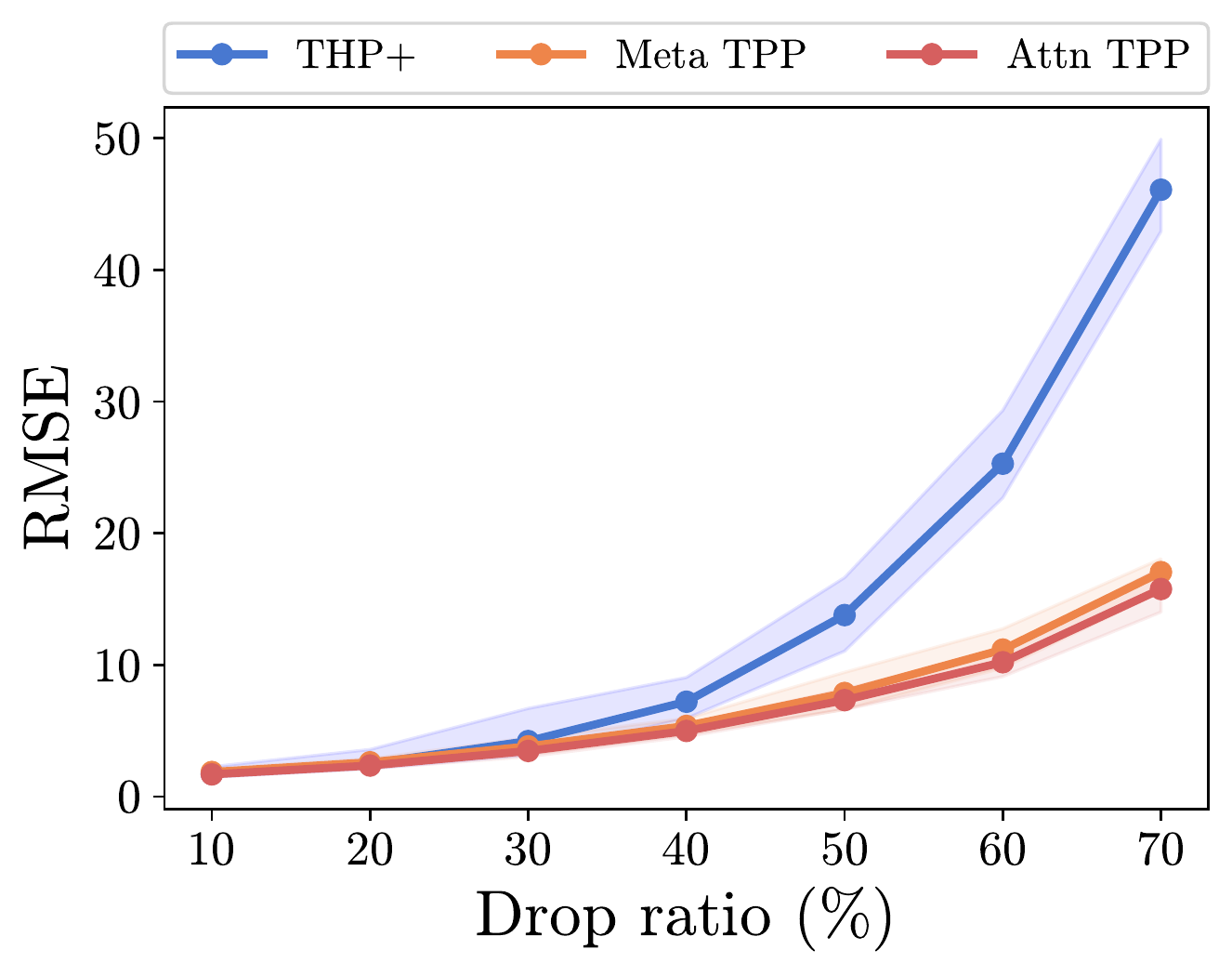}
        \caption{Imputation with different drop rates}
        \label{fig:imputation}
    \end{subfigure}
    \begin{subfigure}[t]{0.48\textwidth}
        \centering
        \includegraphics[width=0.95\textwidth]{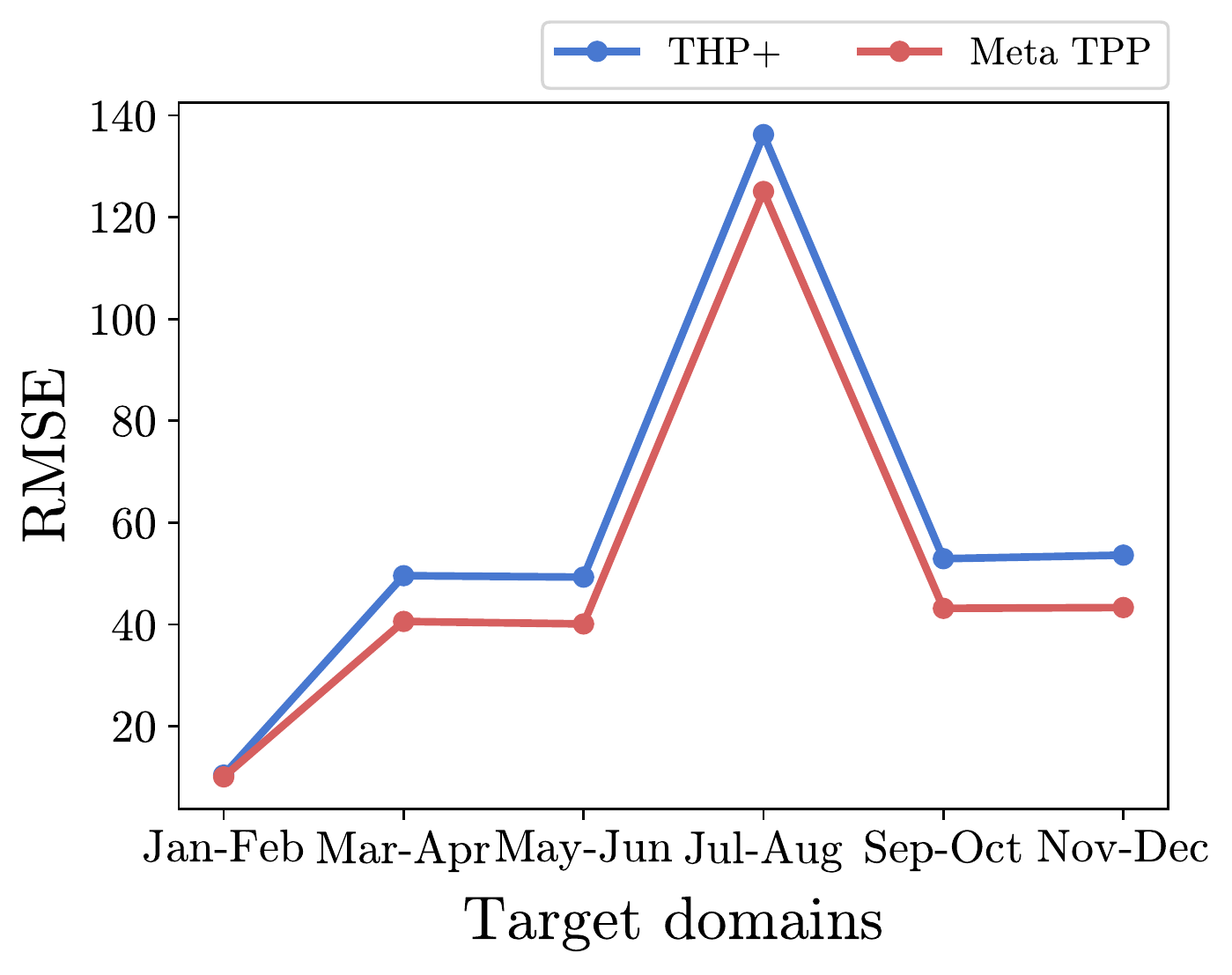}
        \caption{Distribution drift in NYC Taxi dataset}
        \label{fig:domain_shift}
    \end{subfigure}
    \vspace{-1mm}
    \caption{Experiment results on imputation and distribution drift.}
    \vspace{-2mm}
\end{figure*}

\textbf{Hyperparameters.}
We grid-search on every combination of dataset and method for learning rate $\in \{0.01, 0.001, 0.0001, 0.00001 \}$ and weight decay $\in \{ 0.01, 0.001, 0.0001, 0.00001 \}$ for fair comparison.
We bootstrap for $200$ times on test sets to obtain the mean and standard deviation (in parentheses) for the metrics in \cref{fig:imputation} and Table 1--3
following \citet{transformer2021yang}.
All the other hyperparameters are fixed throughout the experiments, and are reported in \cref{app:sec:hyperparameters}.

\subsection{Experiment Results}
\label{subsec:experiment}

In this section, we begin by comparing our attentive meta temporal point process (denoted as Attentive TPP) with state-of-the-art supervised TPP methods on $4$ popular benchmarks. 
We then investigate how Attentive TPP captures periodic patterns, and show how Attentive TPP can be used to impute missing events in noisy sequences. Finally, we consider robustness under distribution drift.

\begin{table*}[t!]
\centering
\setlength{\tabcolsep}{3.7pt}     %
\setlength{\cmidrulekern}{0.25em} %
\fontsize{8.0}{12.0}\selectfont
\begin{tabu}{ccccccccccccc}
\hline
\multirow{2}{*}{Methods} &
\multicolumn{3}{c}{Stack Overflow} &
\multicolumn{3}{c}{Mooc} &
\multicolumn{3}{c}{Reddit} &
\multicolumn{3}{c}{Wiki} \\
\cmidrule(lr){2-4}\cmidrule(lr){5-7}\cmidrule(lr){8-10}\cmidrule(lr){11-13}
& RMSE  & NLL  & Acc & RMSE  & NLL  & Acc & RMSE  & NLL  & Acc & RMSE & NLL  & Acc \\
\hline
\hline
\multirow{2}{*}{Intensity-free}
& 3.64 & 3.66 & 0.43 
& 0.31  & 0.94  & \bf{0.40}  
& 0.18 & 1.09 & 0.60 
& 0.60 & 7.76 & \bf{0.26} \\
& (0.26) & (0.02) & (0.005)
& (0.006) & (0.03) & \bf{(0.004)}
& (0.006) & (0.04) & (0.008)
& (0.05) & (0.40) & \bf{(0.03)} \\
\multirow{2}{*}{Neural flow}
& -- & -- & -- 
& 0.47 & 0.43 & 0.30
& 0.32 & 1.30 & 0.60  
& 0.56 & 11.55 & 0.05 \\
& -- & -- & --
& (0.006) & (0.02) & (0.04)
& (0.04) & (0.33) & (0.07)
& (0.05) & (2.22) & (0.01) \\
\multirow{2}{*}{THP$^+$}
& 1.68 & 3.28 & \bf{0.46}
& 0.18 & 0.13 & 0.38 
& 0.26 & 1.20 & 0.60
& 0.17 & \bf{6.25} & 0.23 \\
& (0.16) & (0.02) & \bf{(0.004)} 
& (0.005) & (0.02) & (0.004)
& (0.005) & (0.04) & (0.007) 
& (0.02) & \bf{(0.39)} & (0.03) \\
\hline
\multirow{2}{*}{Attentive TPP}
&  \bf{1.15} & \bf{2.64} & \bf{0.46}
&  \bf{0.16} & \bf{-0.72}  & 0.36
& \bf{0.11} & \bf{0.03} & \bf{0.60}
& \bf{0.15} & \bf{6.25} & 0.25 \\
& \bf{(0.02)} & \bf{(0.02)} & \bf{(0.004)}
& \bf{(0.004)} & \bf{(0.02)} & (0.003)
& \bf{(0.002)} & \bf{(0.04)} & \bf{(0.007)} 
& \bf{(0.01)} & \bf{(0.38)} & (0.03) \\
\Xhline{2\arrayrulewidth}
\end{tabu}
\caption{Comparison of the Attentive TPP to the state-of-the-art methods on a bootstrapped test sets.}
\label{tbl:sota}
\end{table*}

\textbf{Comparison with state-of-the-art methods.} \cref{tbl:sota} summarizes our comparison of Attentive TPP with state-of-the-art baselines -- intensity-free~\citep{intensity_free2019shchur}, neural flow~\citep{nf2021bilovs}\footnote{Neural flow results on Uber, NYC Taxi and Stack Overflow (in \cref{tbl:sota}) datasets are missing because the official implementation runs into NaN values for long event sequences in inversion step.}, and THP$^+$~\citep{thp2020zuo} on the Stack Overflow, Mooc, Reddit, and Wiki benchmarks. THP$^+$ generally performs better than the intensity-free and neural flow baselines. Attentive TPP further improves over THP$^+$ on all datasets and metrics except for mark accuracy on Mooc and Wiki.

\begin{table*}[t!]
\centering
\setlength{\tabcolsep}{3.7pt}     %
\setlength{\cmidrulekern}{0.25em} %
\fontsize{9.0}{12.0}\selectfont
\begin{tabu}{ccccccc}
\hline
\multirow{2}{*}{Methods} &
\multicolumn{2}{c}{Sinusoidal} &
\multicolumn{2}{c}{Uber} &
\multicolumn{2}{c}{NYC Taxi} \\
\cmidrule(lr){2-3}\cmidrule(lr){4-5}\cmidrule(lr){6-7}
& RMSE  & NLL 
& RMSE  & NLL & RMSE  & NLL \\
\hline
\hline
Intensity-free
& 1.29 (0.08) & 0.88 (0.02) 
& 51.23 (2.89) & 4.46 (0.02)
& 46.59 (26.16) & 2.06 (0.07) \\
Neural flow
& \bf{1.13} (0.07) & 0.99 (0.02)
& -- & -- & -- & -- \\
THP$^+$
& 1.72 (0.10) & 0.84 (0.02)
& 90.25 (4.53) & 3.63 (0.03)  
& 10.31 (0.47) & \bf{2.00} (0.01)  \\
\hline
Attentive TPP (Ours) 
& 1.45 (0.11) & \bf{0.66} (0.02)
& \bf{22.11} (1.94) & \bf{2.89} (0.04) 
& \bf{8.92} (0.42) & \bf{2.00} (0.009) \\
\Xhline{2\arrayrulewidth}
\end{tabu}
\caption{Experiment results on bootstrapped test sets with strong periodic patterns. }
\vspace{-3mm}
\label{tbl:periodic}
\end{table*}

\begin{figure*}[t!]
    \centering
    \begin{subfigure}[t]{0.53\textwidth}
        \centering
        \includegraphics[width=0.92\textwidth]{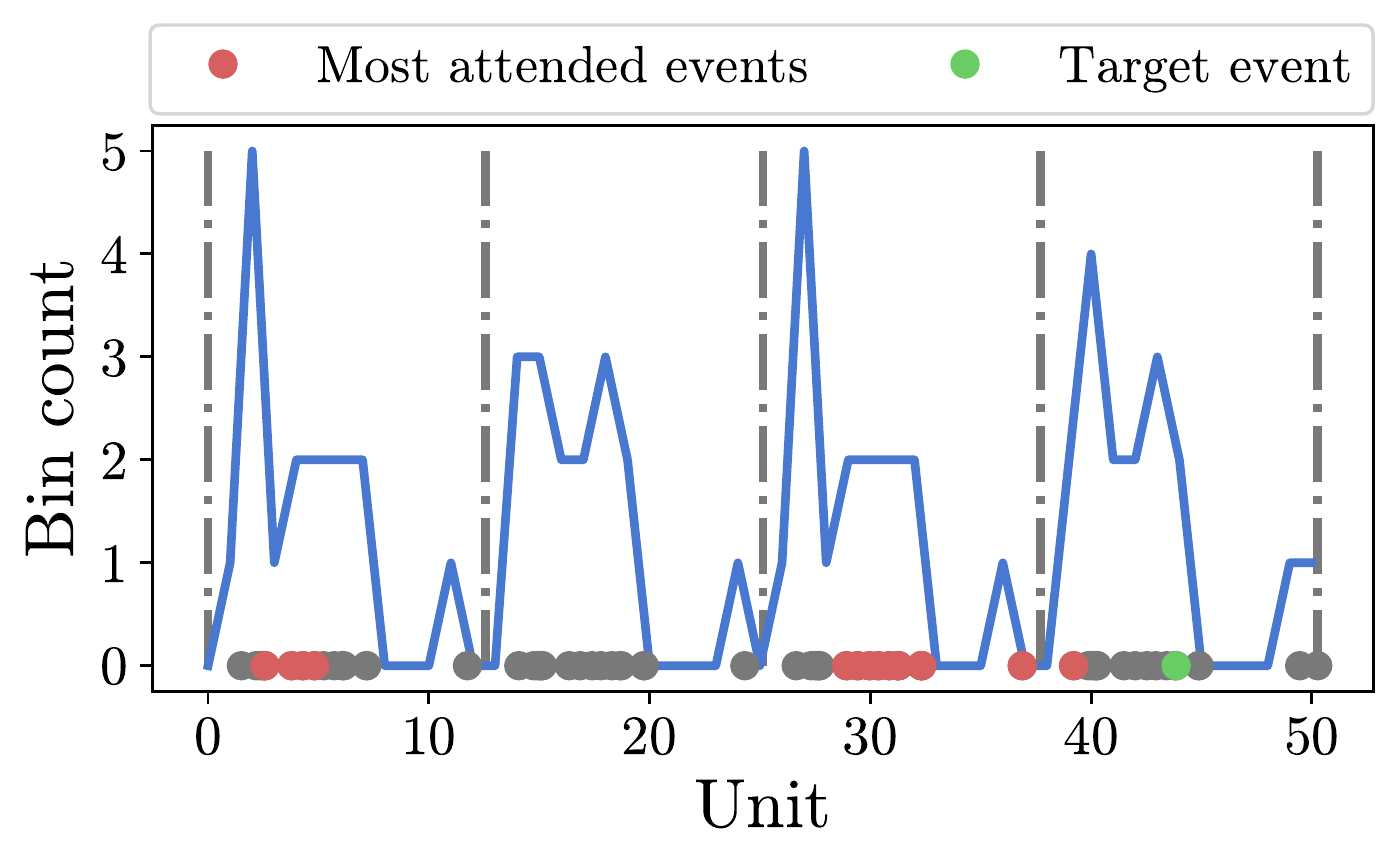}
        \vspace{-1.5mm}
        \caption{Self-attention in THP and proposed cross-attention}
        \label{fig:attention}
    \end{subfigure}%
    ~ 
    \begin{subfigure}[t]{0.44\textwidth}
        \centering
        \includegraphics[width=0.92\textwidth]{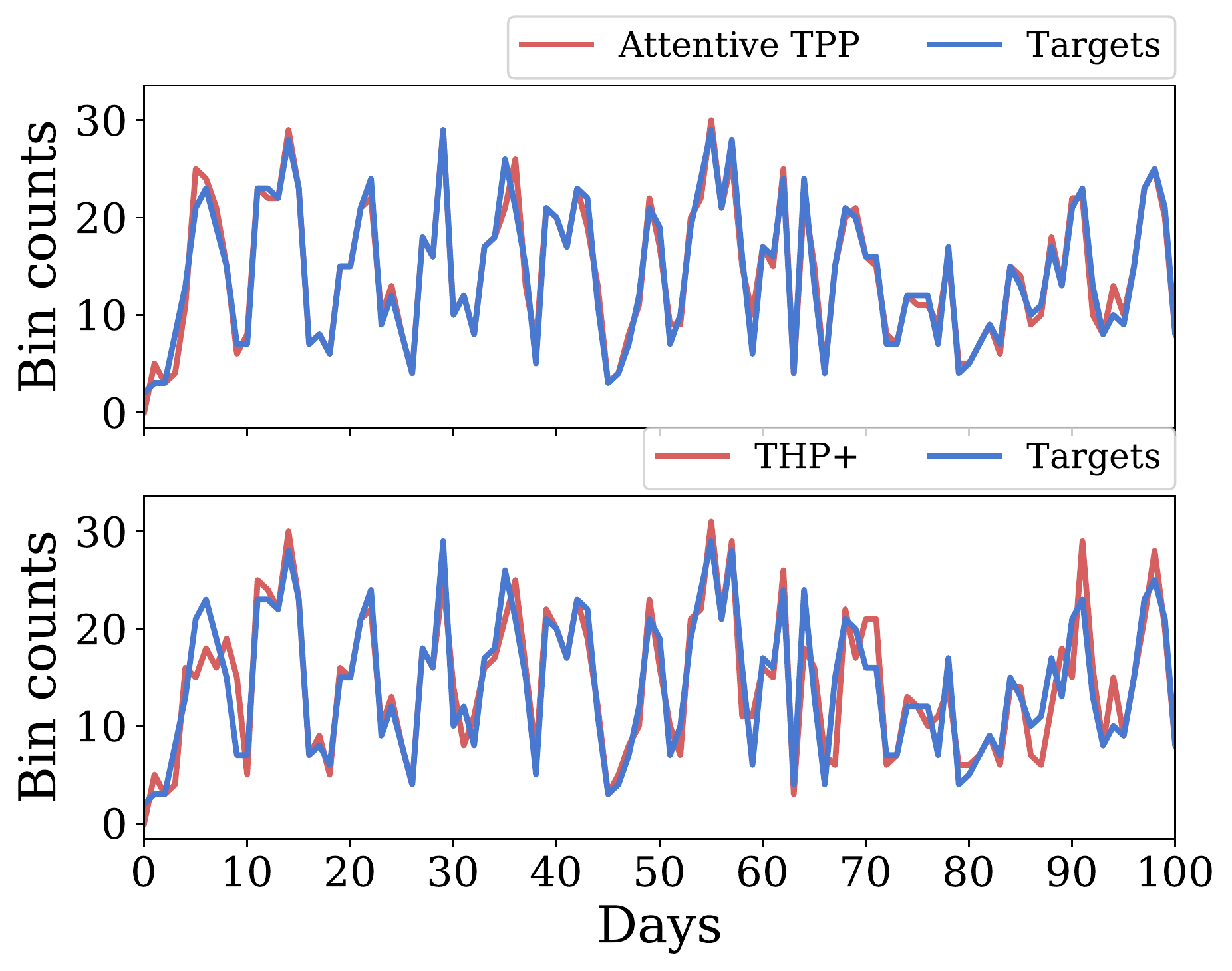}
        \vspace{-1.5mm}
        \caption{Visualization of test predictions vs. targets}
        \label{fig:qualitative}
    \end{subfigure}
    \vspace{-1.5mm}
    \caption{Qualitative analysis on the cross-attention and prediction results.}
    \vspace{-3.5mm}
\end{figure*}

\textbf{Periodic patterns.} 
As previously mentioned in \cref{subsec:attentive_tpp}, the cross-attention module is designed to capture periodic patterns by matching the local history of the current event to the local histories of the previous event times, in addition to alleviating the underfitting problem.
To validate the effectiveness of the cross-attention, we experiment on the datasets with strong periodicity -- Sinusoidal, Uber, and NYC Taxi (please refer to \cref{app:sec:datasets} for details).
\cref{tbl:periodic} shows that the Attentive TPP generally outperforms the state-of-the-art methods, except for RMSE on Sinusoidal.
To investigate the behavior of the cross-attention, we provide an example in \cref{fig:attention} where we highlight 15 (out of 64) the most attended local history indices (in red) to predict the target event (in green) in a sequence from Sinusoidal.
The dotted grey lines represent the start and end of periods.
We can observe that the attention refers to the local histories with similar patterns more than the recent ones.

\subsection{Applications}

\textbf{Imputation.} 
We study the robustness of Meta and Attentive TPP to noise by randomly dropping events, simulating partial observability in a noisy environment, and measuring imputation performance.
For the experiment, we drop $n$ percentage of all the event times drawn independently at random per sequence on the Sinusoidal wave dataset.
In \cref{fig:imputation}, we report the bootstrapped imputation performance of THP$^+$, Meta TPP, and Attentive TPP, in terms of RMSE.
As the drop ratio increases, RMSE increases for all three models but the gap exponentially increases.
Given that the performance gap between three models on `next event' predictions is not as large (mean RMSE -- THP$^+$: $1.72$, Meta TPP: $1.49$, Attentive TPP: $1.45$),
the results shown in \cref{fig:imputation} imply that the Meta and Attentive TPP are significantly more robust to the noise coming from partial observability. %

\textbf{Distribution drift.} 
Distribution drift occurs when the distribution observed during training becomes misaligned with the distribution during deployment due to changes in the underlying patterns over time. This is a common deployment challenge in real-world systems.
\cref{fig:domain_shift} shows how THP$^+$ and Meta TPP models trained on the January-February data of the NYC Taxi dataset generalize to subsequent months. Both models show a decrease in performance, suggesting the presence of non-stationary or seasonal patterns in the data that are not captured in the training months; however, Meta TPP is comparatively more robust across all out-of-domain settings.
It is also worth mentioning that although the Attentive TPP generally performs better than Meta TPP in the conventional experimental setting, it is not the case for distribution drifts.
We conjecture it is because the cross-attention is designed to alleviate the underfitting problem, which results in being less robust to distribution drift.

\begin{table*}[t!]
\begin{minipage}{0.52\textwidth}
\centering
\fontsize{7.5}{12.0}\selectfont
\begin{tabu}{cccccccc}
\hline
\multirow{2}{*}{Attention}
& \multirow{2}{*}{Latent}
& \multicolumn{3}{c}{Reddit}  &\multicolumn{2}{c}{Uber} \\
\cmidrule(lr){3-5}\cmidrule(lr){6-7}
& & RMSE & NLL & Acc & RMSE & NLL  \\
\hline
\hline
\xmark &  \xmark      & 0.16 & 0.92 & 0.59 & 63.71 & 3.68 \\
\xmark &  \cmark  & 0.13 & \bf{-0.39} & \bf{0.61} & 63.35 & 3.25 \\
\cmark  & \xmark   & \bf{0.12} & 0.29 & \bf{0.61} & 47.91 & 3.72 \\
\cmark & \cmark & \bf{0.12} & 0.07 & 0.60 & \bf{21.87} & \bf{2.98}  \\
\Xhline{2\arrayrulewidth}
\end{tabu}
\caption{Comparison of the variants of Meta TPPs}
\label{tbl:meta_variants}
\end{minipage}%
\begin{minipage}{0.53\textwidth}
\centering
\fontsize{7.5}{12.0}\selectfont
\begin{tabu}{ccccc}
\hline
& & 
\multicolumn{3}{c}{Reddit} \\
\cmidrule(lr){3-5}
 Methods & $\#$ Params & RMSE & NLL & Acc \\
 \hline
 \hline
THP$^+$  & 113K & 0.26 & 1.19 & \bf{0.60} \\
 & 170K & 0.29 & 0.79 & 0.59 \\
 & 226K & 0.28 & 1.44 & 0.57 \\
 \hline
 AttnTPP & 222K & \bf{0.12} & \bf{0.07} & \bf{0.60} \\
\Xhline{2\arrayrulewidth}
\end{tabu}
\caption{Comparison of diff. model size}
\label{tbl:model_size}
\end{minipage}%
\vspace{-4mm}
\end{table*}

\subsection{Ablation Studies}
\label{subsec:ablation}

\textbf{Meta TPP and its variants.}
In \cref{tbl:meta_variants}, we compare the proposed Meta TPP and its variants on Reddit, and Uber datasets.
The result shows that both cross-attention and latent variable path generally help to improve the performance. When they are combined (resulting in the Attentive TPP), it generally performs the best in terms of both RMSE and NLL.

\textbf{Different model sizes.}
Both latent path and cross-attention components introduce additional learnable parameters (for the Reddit dataset with $984$ classes, THP$^+$: $113$K, Meta TPP: $126$K, and Attentive TPP: $209$K parameters).
We provide an ablation study with varying number of model parameters for the THP$^+$ baseline to validate the performance improvement does not come from the increased number of model parameters.
In \cref{tbl:model_size}, we increase the number of model parameters for THP$^+$ on Reddit along with the result of the Attentive TPP.
The result shows that the larger model does not necessarily help to improve the performance: as the number of parameters increases, NLL sometimes improves but it may hurt RMSE as in the case of \cref{tbl:model_size}.
The significant improvement in performance of our proposed method shows the importance of providing an effective inductive bias.

\textbf{Parameter sharing.}
In Attentive NP, the encoders of the latent path and attention path are separated to provide different features. 
However, it can significantly increase computational overhead, and for this reason, we share the weights for the encoders.
As an ablation study, we provide the performance with and without sharing the weights for the encoders of the Attentive TPP on the Stack Overflow dataset.
Although the number of parameters of `with sharing' is $50\%$ less than `without sharing' (`without sharing': $86$K vs. `with sharing': $136$K), it performs better than `without sharing' (RMSE / NLL -- `without sharing': $1.08 \,/ \,3.20$ vs. `with sharing': $1.03 \,/\, 2.81$).

\textbf{Visualization of event time predictions.}
In TPP literature, the evaluation relies only on the RMSE and NLL metrics.
It is, however, often hard to measure how practically useful a trained TPP model is.
To qualitatively evaluate TPP models, we convert an event time sequence into time series sequence: we count the number of event times falling into each bin (in \cref{fig:qualitative}, each bin is a day).
\cref{fig:qualitative} shows how close overall predictions of the Attentive TPP and THP$^+$ (in red) are to the ground truth event times (in blue).
In the figure, we can see that the Attentive TPP's predictions closely align with the targets whereas the predictions of the THP$^+$ are off at some regions.
Note that as the y-axis represents bin counts, even a slight off from the ground truth implies large values in terms of RMSE.

\section{Conclusion}
\label{sec:conclusion}

Previously, neural temporal point processes (TPPs) train neural networks in a supervised learning framework.
Although performing well on event sequences similar to a training set, they are susceptible to overfitting, and may not generalize well on sequences with unseen patterns.
To alleviate this, we proposed a novel framework for TPPs using neural processes (NPs).
We proposed the Meta TPP where TPP is formulated as NP, and further developed the Attentive TPP using a cross-attention module, which forces a model to use similar features for repeating local history patterns.
Our experiments demonstrate that the proposed models outperform strong state-of-the-art baselines on several event sequence datasets, effectively capture periodic patterns, and increase robustness to noise and distribution drift.
We believe this work opens up a new research direction to meta learning for TPPs.

\clearpage

\section*{Reproducibility Statement}
Hyperparameters and implementation details are available in \cref{subsec:experiment_setting} and \cref{app:sec:hyperparameters}.
The code for the baselines and the proposed method will be released upon acceptance.
Our code is based on publicly available official \href{https://github.com/shchur/ifl-tpp}{intensity-free} and \href{https://github.com/mbilos/neural-flows-experiments}{neural flow} code.

\bibliography{iclr2023_conference}
\bibliographystyle{iclr2023_conference}

\clearpage
\appendix

\section{Derivation of the Evidence Lower Bound}
\label{app:sec:elbo}

We provide the detailed steps to derive the evidence lower bound of the proposed Meta TPP model shown in \cref{eq:npvi}.
Here, Equation (8) to (9) holds with Jensen's inequality.
\begin{align*}
    \log p_\theta(\tau_l \vbar \tau_{l-k:l-1}, \mathcal{C}_l) &= \log \int p_\theta(\tau_l \vbar \tau_{l-k:l-1}, z) p_\theta(z \vbar \mathcal{C}_l) dz \\
    &= \log \int p_\theta(z | \mathcal{C}_L) \frac{p_\theta(z | \mathcal{C}_l)}{p_\theta(z | \mathcal{C}_L)} p_\theta(\tau_l | \tau_{l-k:l-1}, z) dz \\
    &\geq \int p_\theta(z | \mathcal{C}_L) \log \frac{p_\theta(z | \mathcal{C}_l)}{p_\theta(z | \mathcal{C}_L)} p_\theta(\tau_l | \tau_{l-k:l-1}, z) dz \\
    &= \mathbb{E}_{z \sim p_\theta(z \vbar \mathcal{C}_L)} \left[ \log p_\theta(\tau_l \vbar \tau_{l-k:l-1}, z) \right] - KL(p_\theta(z \vbar \mathcal{C}_L) \vbar p_\theta(z \vbar \mathcal{C}_l)).
\end{align*}

\section{Role of Global Latent Feature} 
\label{app:sec:global_feature}
For the global latent feature $z$ to be task-specific, it has to be distinct for different sequences but similar throughout different event times $l \in [1, L-1]$ within the same sequence as mentioned in \cref{subsec:meta_tpp}.
It is natural for the global features to be distinct by sequence but we need further guidance to make the global feature shared across all the event times in a sequence.
In fact, due to the permutation invariance constraint (implemented in average-pooling), the global latent feature $z$  cannot be very different at different event time: adding some addition context features $r_i$ will not change $G$ as well as $z$ much.

For the latent variable $z$, additional guidance is provided, which is clearer with the objective of the variational inference.
Recall that the objective of the variational inference in Equation (6) is provided as,
$$\argmax_\theta E_{z \sim p_\theta(z \vbar \mathcal{C}_L)}  \log p_\theta(\tau_l \vbar \tau_{l-k:l-1}, z)  - KL(p_\theta (z \vbar \mathcal{C}_L) \;||\; p_\theta (z \vbar \mathcal{C}_l)).$$
Here, regardless of the index of the target $l$, it always minimizes the KL divergence between $p_\theta (z \vbar \mathcal{C}_L)$ and $p_\theta ( z \vbar \mathcal{C}_l)$ where $L$ is the length of a sequence. 
So, ideally, the latent variable $z \sim p_\theta (z \vbar \mathcal{C}_l)$ should capture the same information as $z \sim p_\theta (z \vbar \mathcal{C}_L)$. 
It implies regardless of the index of the target $l$, the latent variable $z$ asymptotically captures the global feature of the whole sequence, which is equivalent to $z \sim p_\theta (z \vbar \mathcal{C}_L)$.
Hence, the resulting $p_\theta (z \vbar \mathcal{C}_l)$ captures the global and task-specific patterns, which ideally is similar to $p_\theta (z \vbar \mathcal{C}_L)$.
As a result, the global latent feature is guided to be distinct for different sequences but similar throughout different event time $l \in [1, L-1]$ within the same sequence.

It is also worth mentioning that its role is quite different from the target input $\tau_{l-k+1:l}$ which is another input for the decoder. 
Consider two events that are far apart from each other.
Due to distinctive local patterns, their target inputs can be quite different from each other. 
On the other hand, the global features will not be that different as they are the average of all the context features at each event time, plus guided by the KL divergence. 
Hence, the global feature provides ``overall’’ patterns of a task whereas a target input provides local patterns to the decoder.

Back to our original goal: treating each sequence as a realization of a distinct stochastic process, we use the global latent feature that is distinct by each sequence to provide a task-specific information which is shared regardless of different event time step $l$.
It is neither implicitly nor explicitly considered in the supervised learning case.
In supervised learning, each event time step at each sequence is treated equally from which patterns for only one stochastic process is learned.

In the \cref{app:tbl:global_feature}, we compare the THP$^+$ baseline and Meta TPP on Sinusoidal, Uber and NYC Taxi datasets to demonstrate the effectiveness of the global latent feature.
The decoder of the Meta TPP takes the global latent feature $z$ (from the permutation invariance constraint) as an input, in addition to the target input feature $r\_l$ that the decoder of the THP$^+$ baseline takes as input.  
The result shows that the global latent feature generally helps to improve both RMSE and NLL performance.

\begin{table*}[t!]
\centering
\begin{tabu}{ccccccc}
\hline
\multirow{2}{*}{Methods}
& \multicolumn{2}{c}{Sinusoidal} 
& \multicolumn{2}{c}{Uber}
& \multicolumn{2}{c}{NYC Taxi} \\
\cmidrule(lr){2-3}\cmidrule(lr){4-5}\cmidrule(lr){6-7}
& RMSE & NLL & RMSE & NLL & RMSE & NLL  \\
\hline
\hline
THP$^+$ &  1.72 & 0.84 & 90.25 & 3.63 & 10.31 & \bf{2.00} \\
Meta TPP & \bf{1.48} & \bf{0.61} & \bf{63.35} & \bf{3.25} & \bf{10.04} & 2.33 \\
\Xhline{2\arrayrulewidth}
\end{tabu}
\caption{Comparison between THP$^+$ baseline and Meta TPP.}
\label{app:tbl:global_feature}
\end{table*}

\section{Computation of Evaluation Metrics}
\label{app:sec:eval}

Unlike Equation (7) in the main paper where the ELBO is computed using samples from $p_\theta(z \vbar \mathcal{C}_L)$, in inference, we do not have access to $z \sim p_\theta(z \vbar \mathcal{C}_L)$.
But, as $p_\theta(z \vbar \mathcal{C}_l)$ is trained to be similar to $p_\theta(z \vbar \mathcal{C}_L)$ through $KL(p_\theta(z \vbar \mathcal{C}_L) \vbar p_\theta(z \vbar \mathcal{C}_l)$, we use samples $z \sim p_\theta(z \vbar \mathcal{C}_l)$.
As specified in \cref{app:sec:hyperparameters}, we use 256 samples to have good enough approximation.  
\begin{itemize}
    \item NLL -- We approximate a log-likelihood of the next event time $\tau_{l+1}$ using Monte-Carlo approximation as, 
    \begin{align}
        \log p_\theta(\tau_{l+1} \vbar \tau_{l-k+1:l}, \mathcal{C}_l)
        &= \log \int p_\theta(\tau_{l+1} \vbar \tau_{l-k+1:l}, z) p_\theta(z \vbar \mathcal{C}_l) dz \\
        &\approx \log \frac{1}{M} \sum_{m=1}^M p_\theta (\tau_{l+1} \vbar \tau_{l-k+1:l}, z_m)
    \end{align}
    where $M$ is the number of samples from $p_\theta(z \vbar \mathcal{C}_l)$.
    \item RMSE -- 
    We use a mixture of log-normal distributions to model $p_\theta(\tau_{l+1} \vbar \tau_{l-k+1:l}, z)$.
    Formally, for $l \in [1, L-1]$, $\tau_{l+1} \sim MixLogNorm(\bm{\mu}_{l+1}, \bm{\sigma}_{l+1}, \bm{\omega}_{l+1})$ where $\bm{\mu}_{l+1}$ are the mixture means, $\bm{\sigma}_{l+1}$ are the standard deviations, and $\bm{\omega}_{l+1}$ are the mixture weights.
    The parameters are the outputs of the decoder given a latent sample $z$.
    Knowing this, we can analytically compute the expected event time for a latent sample $z$ with $K$ mixture components as, 
    $$E_{\tau_{l+1} \sim p_\theta (\tau_{l+1} \vbar \tau_{l-k+1:l}, z)} [\tau_{l+1}] = \sum_{k=1}^K \omega_{l+1,k} \exp{(\mu_{l+1,k} + \frac{1}{2} \sigma_{l+1, k}^2)}.$$ 
    Note that since this expectation is over $p_\theta (\tau_{l+1} \vbar \tau_{l-k+1:l}, z)$ where $z$ is one sample from the posterior, we need to take another expectation over the posterior as follows,
    \begin{align}
        E_{\tau_{l+1} \sim p_\theta (\tau_{l+1} \vbar \tau_{l-k+1:l}, \mathcal{C}_l)} [ \tau_{l+1} ]
        &= E_{z \sim p_\theta (z \vbar \mathcal{C}_l)}  E_{\tau_{l+1} \sim p_\theta (\tau_{l+1} \vbar \tau_{l-k+1:l}, z)} [ \tau_{l+1} ] \\
        &= E_{z \sim p_\theta (z \vbar \mathcal{C}_l)} \sum_{k=1}^K \omega_{l+1,k} \exp{(\mu_{l+1,k} + \frac{1}{2} \sigma_{l+1, k}^2)} \\
        &\approx \frac{1}{M} \sum_{m=1}^M \sum_{k=1}^K \omega_{l+1,k} \exp{(\mu_{l+1,k} + \frac{1}{2} \sigma_{l+1, k}^2)}
    \end{align}
    where $M$ is the number of samples from $p_\theta(z \vbar \mathcal{C}_l)$.
    \item Accuracy -- We obtain class predictions by taking argmax over the probability distribution of class labels as follows,   
    $$\argmax_{c \in [1, C]} p_\theta (y_{l+1} \vbar \tau_{l-k+1:l}, y_{l-k+1:l}, \mathcal{C}_l)$$
    where $C$ is the number of marks.
    The probability distribution of class labels is approximated using MC samples as,
    \begin{align}
        p_\theta (y_{l+1} \vbar \tau_{l-k+1:l}, y_{l-k+1:l}, \mathcal{C}_l) &= \int p_\theta (y_{l+1} \vbar r_l, z) p_\theta (z \vbar \mathcal{C}_l) dz \\
        &\approx \frac{1}{M} \sum_{m=1}^M p_\theta (y_{l+1} \vbar r_l, z_m)
    \end{align}
    where $M$ is the number of samples from $p_\theta(z \vbar \mathcal{C}_l)$.
\end{itemize}

\section{Monte-Carlo Approximation vs. Variational Inference}
\label{app:sec:mc_vs_vi}

Amortized variational inference (VI) we described in \cref{sec:meta_tpp} is not the only way to approximate the latent variable model.
We can also use Monte-Carlo (MC) approximation, which is simpler and does not rely on a proxy like ELBO.
It is formulated as,
\begin{align}
    \log \int p_\theta(\tau_{l+1} \vbar \tau_{l-k+1:l}, z) p_\theta(z \vbar \mathcal{C}_l) dz \approx \log \frac{1}{N}\sum_{n=1}^N p_\theta(\tau_{l+1} \vbar \tau_{l-k+1:l}, z_n)
     \label{eq:npml}
\end{align}
Note that a sample $z_n$ in \cref{eq:npml} is drawn from $p(z_n | \mathcal{C}_l)$, which is different from $z_n \sim p(z_n | \mathcal{C}_L)$ in \cref{eq:npvi}. 
\citet{npml2020foong,npf2020dubois} report that MC approximation generally outperforms variational inference.
In variational inference, as a model is trained with $z \sim p_\theta(z \vbar \mathcal{C}_L)$, the samples $z \sim p_\theta(z \vbar \mathcal{C}_l)$ in test time are quite different from what the model has used as inputs: although $KL(p_\theta(z \vbar \mathcal{C}_L) \vbar p_\theta(z \vbar \mathcal{C}_l))$ forces $p_\theta(z \vbar \mathcal{C}_L)$ and $p_\theta(z \vbar \mathcal{C}_l)$ close to each other, it is hard to make KL-divergence to be zero.
Although MC approximation outperforms VI for the proposed Meta TPP, it is not the case for the Attentive TPP as shown in \cref{tbl:mc_vs_vi}. 

\begin{table*}[h!]
\begin{minipage}{0.98\textwidth}
\centering
\fontsize{9.0}{12.0}\selectfont
\begin{tabu}{cccccccccc}
\hline
\multirow{2}{*}{Attention}
& \multicolumn{2}{c}{Latent}
& \multicolumn{3}{c}{Reddit}
& \multicolumn{2}{c}{Uber} 
& \multicolumn{2}{c}{NYC Taxi}\\
\cmidrule(lr){2-3}\cmidrule(lr){4-6}\cmidrule(lr){7-8}\cmidrule(lr){9-10}
& VI & MC & RMSE & NLL & Acc & RMSE & NLL & RMSE & NLL  \\
\hline
\hline
\xmark &  \cmark   & \xmark & 0.13 & \bf{-0.39} & \bf{0.61} & 63.35 & 3.25 & \bf{10.04} & 2.33 \\
\xmark & \xmark & \cmark & \bf{0.11} & 0.16 & \bf{0.61} & \bf{37.12} & \bf{3.22} & 10.15 & \bf{2.00}  \\
 \hline
\cmark & \cmark & \xmark & \bf{0.11} & 0.03 & \bf{0.60} & \bf{21.87} & \bf{2.98} & \bf{8.92} & \bf{2.00} \\
 \cmark & \xmark & \cmark & 0.13 & \bf{-0.05}  & \bf{0.60} & 22.38 & 3.18 & 9.10 & 2.01  \\
\Xhline{2\arrayrulewidth}
\end{tabu}
\caption{Comparison of the variants of Meta TPPs}
\label{tbl:mc_vs_vi}
\end{minipage}%
\end{table*}

We conjecture it based on MC approximation better sharing role with the cross-attention path when compared to the VI approximation.
More specifically, cross-attention forces a model to have similar features for repeating local history patterns.
As it focuses on extracting features from the previous history, which is similar to what $z \sim p_\theta(z \vbar \mathcal{C}_l)$ contains, the latent and attentive path share a role in MC approximation.
On the other hand, as the model is trained on the global latent feature $z \sim p_\theta(z \vbar \mathcal{C}_L)$ in VI, without focusing too much on the previous history due to the cross-attention, it may be able to utilize more diverse features. 
It would be an interesting future work to investigate the theoretical relationship between approximation methods and variants of Meta TPP.

\section{Na\"ive Baseline}
\label{app:sec:naive}
Sometimes na\"ive baselines can be stronger baselines than more sophisticated ones. 
To investigate if that is the case for TPPs, we implement a na\"ive baseline that makes predictions based on median inter-event interval: $\hat{\tau}\_{l+1} = \tau\_{l} + \Delta \tau\_{median, l}$ where $\Delta \tau\_{median,l}$ is a median of the inter-event interval up to $l$-th event. 
We boostrap for 200 times on the test set to obtain the mean of RMSE metrics as with how we obtain the numbers for the other methods (NLLs are not available for the na\"ive baseline).
In \cref{app:tbl:naive}, the performance of the na\"ive baseline is surprisingly good for some cases.
For instance, it is better than the intensity-free on Wiki and NYC Taxi datasets.
It is, however, much worse than THP$^+$ and the proposed Attentive TPP on all the datasets.

\begin{table*}[h!]
\centering
\begin{tabu}{cccccccc}
\hline
Methods
& Stack Overflow 
& Mooc
& Reddit
& Wiki
& Sinusoidal
& Uber
& NYC Taxi \\
\hline
\hline
Na\"ive baseline &   161.21  &  0.79   &  0.38   &  0.21 &  4.61  & 107.91 &  24.58  \\
Intensity-free &  3.64  &  0.31 &  0.18  &  0.60  &   1.29  &   51.23   &  46.59  \\
Neural flow   &   --    &  0.47   &  0.32   &  0.56   & \bf{1.13}   &  --   &   --  \\
THP$^+$ &  1.68  &  0.18 &  0.26  & 0.17 &  1.72  &  90.25  &  10.31  \\
 Attentive TPP  &  \bf{1.15} &  \bf{0.16}   &  \bf{0.11}  &  \bf{0.15}    &  1.45   &  \bf{22.11}   &  \bf{8.92} \\
\Xhline{2\arrayrulewidth}
\end{tabu}
\caption{Comparison of Na\"ive baseline and other methods}
\label{app:tbl:naive}
\end{table*}

\section{Effect of Model Size}
\label{app:sec:model_size}
Although we have demonstrated that the improvement in performance does not come from the size of a model through Table 4 on the Reddit dataset, we provide more evidence on the rest of the datasets as below.

\begin{table*}[h!]
\fontsize{8.0}{10.5}\selectfont
\begin{tabu}{cccccccc}
\multirow{2}{*}{Methods} & \multirow{2}{*}{$\#$ Params} 
& \multicolumn{2}{c}{Sinusoidal} 
& \multicolumn{2}{c}{Uber} 
& \multicolumn{2}{c}{NYC Taxi} \\
\cmidrule(lr){3-4}
\cmidrule(lr){5-6}
\cmidrule(lr){7-8}
&  & RMSE & NLL & RMSE & NLL  & RMSE & NLL  \\
 \hline
 \hline
 THP$^+$  & 50K  & 1.72 (0.10) & 0.84 (0.02) & 90.25 (4.53) & 3.63 (0.03) & 10.31 (0.47) & 2.00 (0.01) \\
 &  100K  & 1.84 (0.13) & 1.04 (0.02) & 82.69 (4.56) & 3.34 (0.03) & 10.16 (0.47) & \bf{1.92 (0.01)} \\
Attentive TPP & 96K  & \bf{1.45 (0.11)} & \bf{0.66 (0.02)} & \bf{22.11 (1.94)} & \bf{2.89 (0.04)} & \bf{8.92 (0.42)} & 2.00 (0.009) \\
 \hline
\Xhline{2\arrayrulewidth}
\end{tabu}
\caption{Comparison of different model size on periodic datasets.}
\label{app:tbl:model_size_periodic}
\end{table*}

\begin{table*}[h!]
\centering
\begin{tabu}{ccccc}
\multirow{2}{*}{Methods} & \multirow{2}{*}{$\#$ Params} 
& \multicolumn{3}{c}{Stack Overflow} \\
\cmidrule(lr){3-5}
&  & RMSE & NLL & Acc  \\
 \hline
 \hline
THP$^+$ &  52K  & 1.68 (0.16) & 3.28 (0.02) & \bf{0.46 (0.004)} \\ 
 & 103K & 1.63 (0.06) &  2.82 (0.03) & \bf{0.46 (0.004)} \\ 
 Attentive TPP &  99K & \bf{1.15 (0.02)} & \bf{2.64 (0.02)} & \bf{0.46 (0.004)}\\
\Xhline{2\arrayrulewidth}
\end{tabu}
\caption{Comparison of different model size on the Stack Overflow dataset.}
\label{app:tbl:model_size_so}
\end{table*}

\begin{table*}[h!]
\centering
\begin{tabu}{ccccc}
\multirow{2}{*}{Methods} & \multirow{2}{*}{$\#$ Params} 
& \multicolumn{3}{c}{Mooc} \\
\cmidrule(lr){3-5}
&  & RMSE & NLL & Acc  \\
 \hline
 \hline
THP$^+$  & 56K  & 0.18 (0.005)  & 0.13 (0.02)  & 0.38 (0.004) \\
        & 113K & 0.22 (0.007)  & 0.05 (0.03)  & \bf{0.39 (0.004)} \\ 
Attentive TPP & 108K & \bf{0.16 (0.004)} & \bf{-0.72 (0.02)} &  0.36 (0.003) \\
\Xhline{2\arrayrulewidth}
\end{tabu}
\caption{Comparison of different model size on the Mooc dataset.}
\label{app:tbl:model_size_mooc}
\end{table*}

\begin{table*}[h!]
\centering
\begin{tabu}{ccccc}
\multirow{2}{*}{Methods} & \multirow{2}{*}{$\#$ Params} 
& \multicolumn{3}{c}{Wiki} \\
\cmidrule(lr){3-5}
&  & RMSE & NLL & Acc  \\
 \hline
 \hline
THP$^+$ & 577K  & 0.17 (0.02) & \bf{6.25 (0.39)}  & 0.23 (0.03) \\ 
        & 1153K  & 0.16 (0.01) & 6.47 (0.40) & 0.21 (0.02) \\ 
Attentive TPP & 1149K  & \bf{0.15 (0.01)} & \bf{6.25 (0.38)}  & \bf{0.25 (0.03)} \\
\Xhline{2\arrayrulewidth}
\end{tabu}
\caption{Comparison of different model size on the Wiki dataset.}
\label{app:tbl:model_size_wiki}
\end{table*}

In the table above, we observe that in many cases, smaller models perform better than larger models : on Sinusoidal, Mooc, and Wiki.
Although it is sometimes true that larger models perform better than their smaller counterparts, they are still significantly worse than our proposed Attentive TPP.
Note that we conducted exactly the same grid search for hyperparameter tuning for the larger models.
The results empirically demonstrate that the improvement does not necessarily come from the size of a model but from right inductive biases.

Lastly, please note that all the experiment results we have reported in Table 1-4 and in rebuttal are on the test sets.
We believe the RMSE, NLL, and Accuracy on test sets are good metrics to compare the generalization performance of different models.
Given that our proposed method outperforms all the baselines, we think our experiments empirically demonstrate the robustness of our method in terms of generalization.

\section{Extension to Marked TPPs}
\label{app:sec:marked_tpp}
We extended the proposed method to the marked cases by adding class prediction branch following the intensity-free TPP (Shchur et al., 2020).
Suppose a mark at $l+1$-th event is denoted as $y_{l+1}$.
For the proposed Meta TPP, we compute the log-likelihood of the mark as,
$$ \log p_\theta (y_{l+1} \vbar \tau_{l-k+1:l}, y_{l-k+1:l}, \mathcal{C}_l) = \log \int p_\theta (y_{l+1} \vbar \tau_{l-k+1:l}, y_{l-k+1:l}, z) p_\theta (z \vbar \mathcal{C}_l) dz. $$
Note that $\mathcal{C}_l$ includes both event times and corresponding labels.
For implementation, we added one fully connected layer that takes as input the same features for the decoder (that predicts the next event time), and outputs the logits for classification.
A class prediction is made by taking argmax over the probability distribution which is approximated using Monte-Carlo samples as,
$$ p_\theta (y_{l+1} \vbar \tau_{l-k+1:l}, y_{l-k+1:l}, \mathcal{C}_l) \approx \frac{1}{M} \sum_{m=1}^M p_\theta (y_{l+1} \vbar r_l, z_m)$$
Note that inputs $\tau_{l-k+1:l}$ and $y_{l-k+1:l}$ are encoded to $r_l$.
The class predictions to compute the accuracies reported throughout the experiments are made from this.

\section{Description of Datasets}
\label{app:sec:datasets}

We use 4 popular benchmark datasets: Stack Overflow, Mooc, Reddit, and Wiki, and 3 newly processed datasets: Sinusoidal wave, Uber and NYC Taxi.
The statistics are provided in \cref{tbl:datasets}.
To split the data into train, validation, and test, we follow the splits made in the previous works such as Shchur et al. (2020), and Yang et al. (2022).
More specifically, we use 60\%, 20\%, and 20\% split for train, validation, and test, respectively, for all the datasets following Shchur et al. (2020) except for Stack Overflow.
For Stack Overflow, we follow the split made by Yang et al. (2022) and Du et al. (2016) where 4,777, 530, and 1,326 samples are assigned for train, validation, and test, respectively.
For more detailed descriptions of the popular benchmark datasets, please refer to the original papers as described below or Section E.2 of \citet{intensity_free2019shchur}.

\begin{table*}[t!]
\centering
\fontsize{9.0}{12.0}\selectfont
\begin{tabu}{ccccc}
\hline
Datasets & $\#$ of Seq. & $\#$ of Events & Max Seq. Length & $\#$ of Marks \\
\hline
\hline
Stack Overflow & 6,633 & 480,414 & 736 & 22 \\
Mooc & 7.047 & 389,407 & 200 & 97 \\
Reddit & 10,000 & 532,026 & 100 & 984 \\
Wiki & 1,000 & 138,705 & 250 & 7,628 \\
\hline
Sinusoidal & 1,000 & 107,454 & 200 & 1 \\
Uber & 791 & 701,579 & 2,977 & 1 \\
NYC Taxi & 1,000 & 1,141,379 & 1,958 & 1 \\
\Xhline{2\arrayrulewidth}
\end{tabu}
\caption{Statistics of the datasets.}
\label{tbl:datasets}
\end{table*}

\subsection{Benchmark Datasets}

\textbf{Stack Overflow.}
It was first processed in \citep{recurrent2016du}.
We use the first folder of the dataset following \citep{intensity_free2019shchur,transformer2021yang}.

\textbf{Mooc.}
It consists of $7{,}047$ sequences, each of which contains of action times an individual user of an online Mooc course.
There are 98 categories.

\textbf{Reddit.}
It consists of $10{,}000$ sequences from the most active users with marks being the sub-reddit categories of each sequence.

\textbf{Wiki.}
It consists of $1{,}000$ sequences from the most edited Wikipedia pages (for a month period) with marks being users who made at least 5 changes.

\subsection{Proposed Datasets}

\textbf{Sinusoidal wave.}
We generate the Sinusoidal wave using a sine function with a periodicity of $4\pi$ and the domain of $[0, 32\pi]$.
We randomly choose the number of events per sequence in $[20, 200]$ for $1{,}000$ sequences.

\textbf{Uber.\footnote{https://www.kaggle.com/datasets/fivethirtyeight/uber-pickups-in-new-york-city/metadata}}
We generate the Uber dataset using the data from the shared link.
Among the data from Januaray, 2015 to June, 2015, we create sequences using \textit{Dispatching-base-num} and \textit{locationID} as keys.
We also give a constraint that the mininum and maximum events per sequence being 100 and $3{,}000$, respectively, and drop all the overlapping event times.

\textbf{NYC Taxi.\footnote{http://www.andresmh.com/nyctaxitrips/}}
We generate the NYC Taxi dataset from the NYC Taxi pickup raw data in 2013 shared in the link, which is different from the one proposed in \citet{recurrent2016du}, as we do not include any location information.
We generate 6 different datasets by splitting the whole data in 2013 for every two months: Jan-Feb, Mar-Apr, May-Jun, Jul-Aug, Sep-Oct, and Nov-Dec.
Throughout the experiment, we train models on the training set of Jan-Feb split, and evaluate on the test set of Jan-Feb for in-domin, and other for distribution drift.

\begin{figure*}[t!]
    \centering
    \begin{subfigure}[t]{1.0\textwidth}
        \centering
        \includegraphics[width=1.0\textwidth]{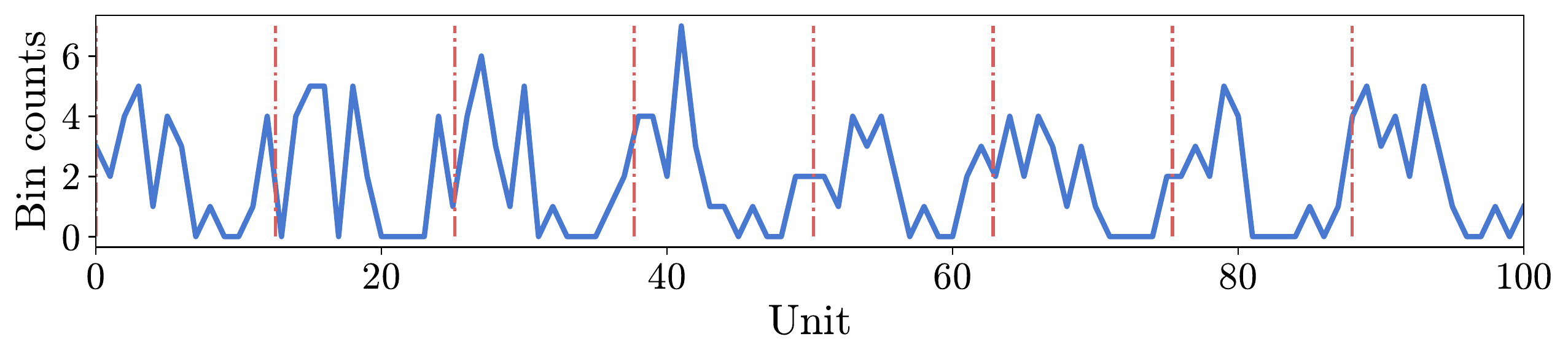}
        \caption{An example in Sinusoidal wave dataset.}
        \label{app:fig:sin_vis}
    \end{subfigure}%
    \qquad
    \begin{subfigure}[t]{1.0\textwidth}
        \centering
        \includegraphics[width=1.0\textwidth]{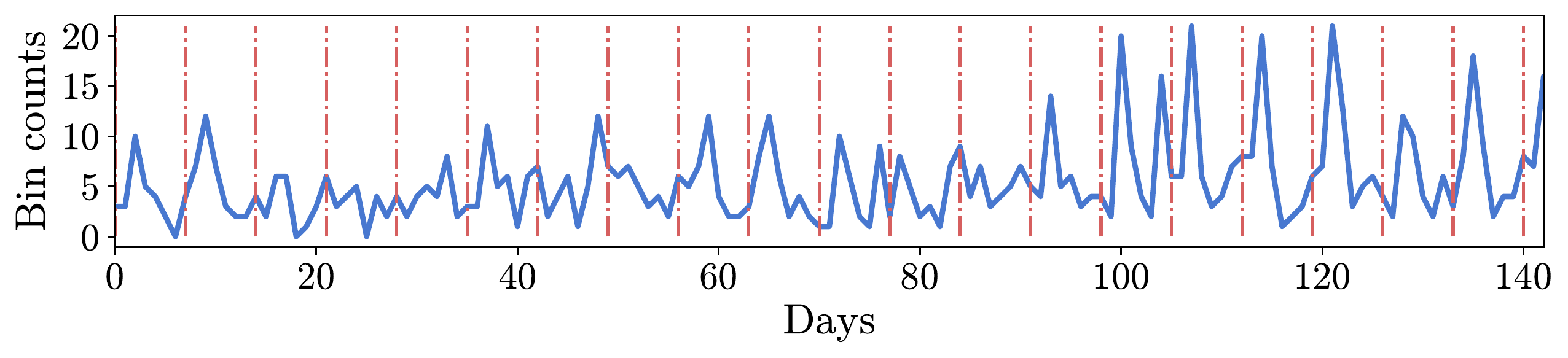}
        \caption{An example in Uber dataset. }
        \label{app:fig:uber_vis}
    \end{subfigure}
    \qquad
    \begin{subfigure}[t]{1.0\textwidth}
        \centering
        \includegraphics[width=1.0\textwidth]{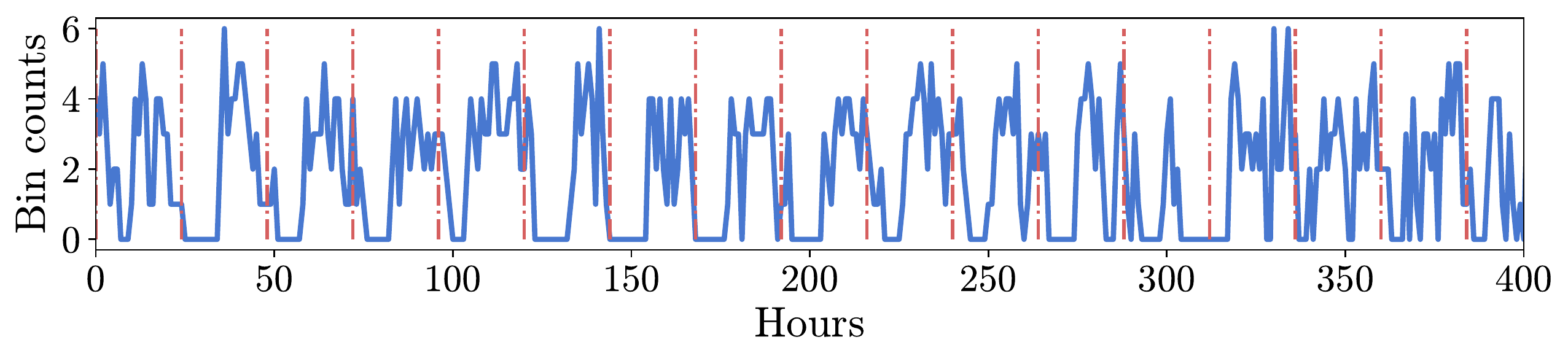}
        \caption{An example in NYC Taxi dataset.}
        \label{app:fig:taxi_vis}
    \end{subfigure}
    \caption{Visualization of examples in the datasets with strong periodicity.}
    \label{app:fig:periodic}
\end{figure*}

\subsection{Periodicity}
In \cref{subsec:experiment}, we provide experiment results that demonstrate the proposed Attentive TPP capture periodic patterns better than the baselines.
To validate that the datasets used for \cref{tbl:periodic} have strong periodic patterns, we provide visualization in \cref{app:fig:periodic}.
Sinusoidal wave dataset has a periodicity for every $4\pi$ as shown in \cref{app:fig:sin_vis}, Uber datset has weekly periodic pattern shown in \cref{app:fig:uber_vis}, and NYC Taxi dataset has daily pattern as shown in \cref{app:fig:taxi_vis}.

\section{Hyperparameters}
\label{app:sec:hyperparameters}
We use the feature dimension of 96, 72, and 64 for the intensity-free~\citep{intensity_free2019shchur}, neural flow~\citep{nf2021bilovs}, and THP$^+$, respectively, as the numbers of parameters fall in the range of $50$K and $60$K with those dimensions.

For the Meta TPP, we use 64 for the dimension of the latent variable $z$, and $32$ samples to approximate the ELBO for variantional inference.
As the variance of variational inference is generally low, $32$ samples are enough to have stable results.
In inference, we increase the sample size to $256$ to have more accurate approximation.
For the Attentive TPP, we use 1-layer self-attention for the cross-attention path, and fix the local history window size to $20$.

For training, we use a batch size of $16$ throughout all the models, and optimize with an Adam optimizer for a grid search for learning rate and weight decay described in \cref{sec:experiments}.

\end{document}